\documentclass[letterpaper]{article} 
\usepackage{titling}
\usepackage{aaai2026}  
\usepackage{times}  
\usepackage{helvet}  
\usepackage{courier}  
\usepackage[hyphens]{url}  
\usepackage{graphicx} 
\usepackage{amsmath}
\usepackage{natbib}  
\usepackage{caption} 
\usepackage{algorithm}
\usepackage{algorithmic}

\usepackage{booktabs}
\usepackage{multirow}
\usepackage{siunitx}
\usepackage{array}
\usepackage{amssymb}
\usepackage{newfloat}
\usepackage{listings}
\DeclareCaptionStyle{ruled}{labelfont=normalfont,labelsep=colon,strut=off} 
\floatstyle{ruled}
\newfloat{listing}{tb}{lst}{}
\floatname{listing}{Listing}
\title{GMF-Drive: Gated Mamba Fusion with Spatial-Aware BEV Representation for End-to-End Autonomous Driving}
\author{
    Jian Wang\textsuperscript{1}\thanks{Equal contribution},
    Chaokang Jiang\textsuperscript{2}\thanks{Equal contribution},
    Haitao Xu\textsuperscript{1}
}
\affiliations{
    \textsuperscript{\rm 1}University of Science and Technology of China\\
    \textsuperscript{\rm 2}China University of Mining and Technology\\
    \{sa20wj@mail.ustc.edu.cn, ts20060079a31@cumt.edu.cn, ba24168190@mail.ustc.edu.cn\}
}

\usepackage{bibentry}

\begin{document}
\maketitle

\begin{abstract}

Diffusion-based models are redefining the state-of-the-art in end-to-end autonomous driving, yet their performance is increasingly hampered by a reliance on transformer-based fusion. These architectures face fundamental limitations: quadratic computational complexity restricts the use of high-resolution features, and a lack of spatial priors prevents them from effectively modeling the inherent structure of Bird's Eye View (BEV) representations. This paper introduces GMF-Drive (Gated Mamba Fusion for Driving), an end-to-end framework that overcomes these challenges through two principled innovations. First, we supersede the information-limited histogram-based LiDAR representation with a geometrically-augmented pillar format encoding shape descriptors and statistical features, preserving critical 3D geometric details. Second, we propose a novel hierarchical gated mamba fusion (GM-Fusion) architecture that substitutes an expensive transformer with a highly efficient, spatially-aware state-space model (SSM). Our core BEV-SSM leverages directional sequencing and adaptive fusion mechanisms to capture long-range dependencies with linear complexity, while explicitly respecting the unique spatial properties of the driving scene. Extensive experiments on the challenging NAVSIM benchmark demonstrate that GMF-Drive achieves a new state-of-the-art performance, significantly outperforming DiffusionDrive. Comprehensive ablation studies validate the efficacy of each component, demonstrating that task-specific SSMs can surpass a general-purpose transformer in both performance and efficiency for autonomous driving.
\begin{figure}[!ht]
\centering
\includegraphics[width=\columnwidth]{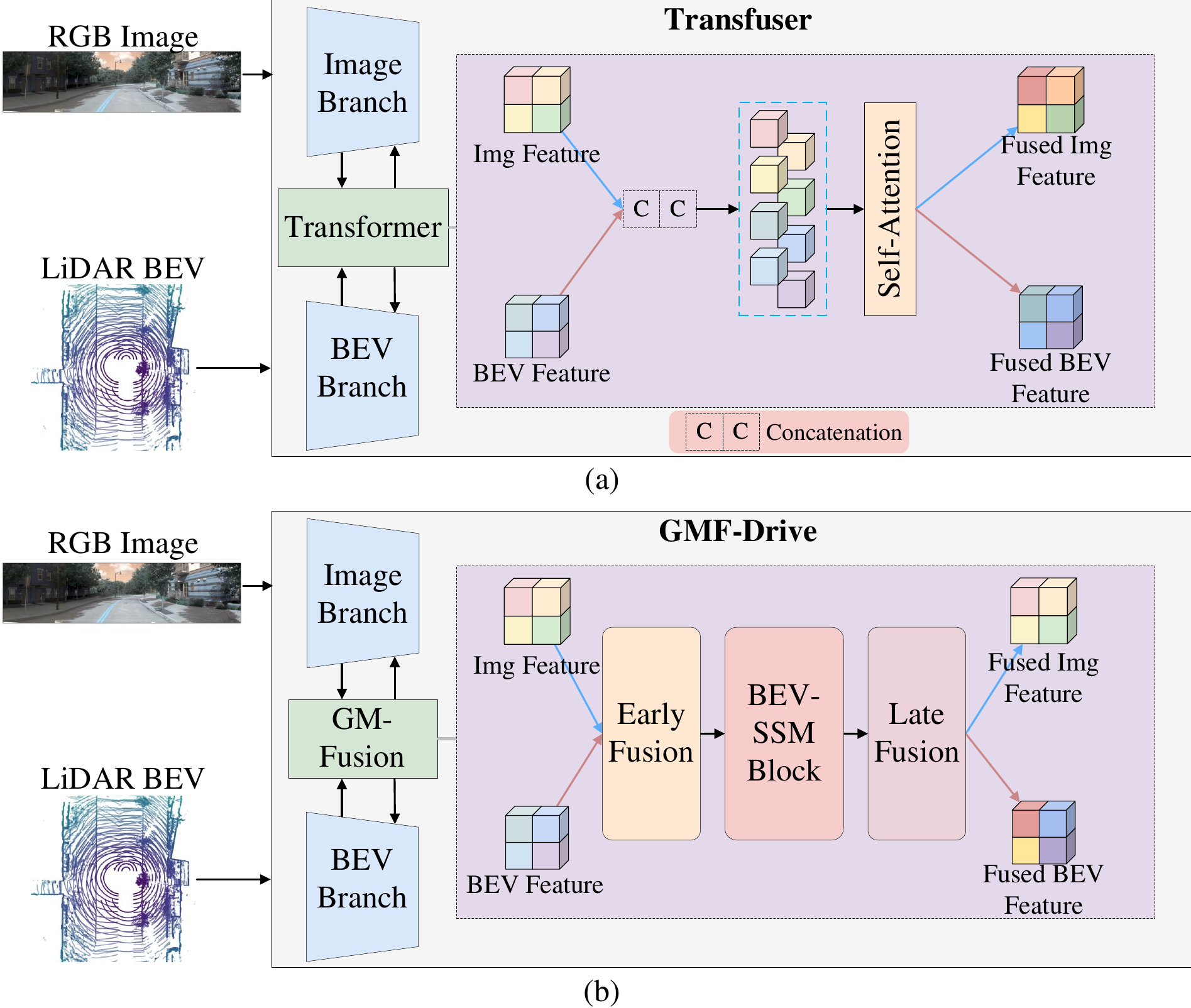} 
\caption{Comparison between the fusion architectures of TransFuser and GMF-Drive. Figure (a) illustrates TransFuser's simple concatenation-based fusion strategy, where image and BEV features are directly concatenated before being processed by self-attention mechanisms. In contrast, Figure (b) shows GMF-Drive's more sophisticated three-stage fusion framework, which first aligns cross-modal features through an early fusion module, then performs efficient spatial modeling via a linear-complexity BEV-SSM Block, and finally achieves deep multimodal integration through a late fusion module, enabling more comprehensive feature fusion than TransFuser's approach.}
\label{Fig. 1}
\end{figure}
\end{abstract}


\begin{figure*}[t]
\centering
\includegraphics[width=\textwidth]{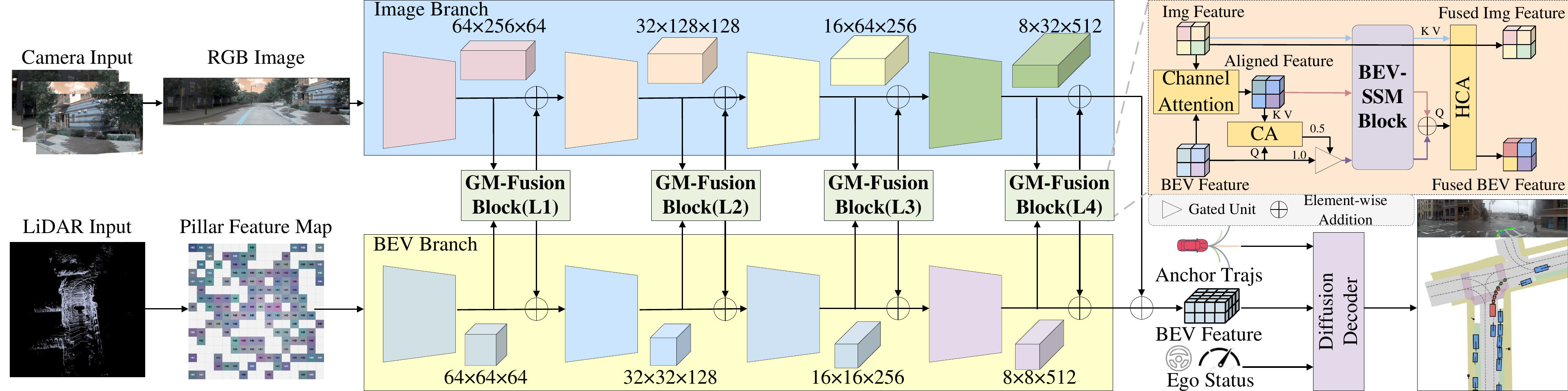} 
\caption{Overview of the GMF-Drive architecture. Left: Multi-modal inputs are processed through separate ResNet-34 backbones with hierarchical GM-Fusion at four scales. Right: Detailed GM-Fusion module showing the information flow - aligned features from channel attention serve as K, V for cross-attention (CA) with BEV queries, followed by gated fusion. Three parallel BEV-SSM blocks process different feature types before final integration via hierarchical deformable \cite{zhu2020deformable} cross-attention (HCA), where image features provide K, V and summed aligned features and BEV features form queries.}
\label{2}
\end{figure*}
\section{Introduction}
End-to-end autonomous driving has emerged as a key approach that directly maps raw sensor inputs to driving actions, reducing reliance on manually designed intermediate representations. Recent developments, particularly diffusion-based planning models like DiffusionDrive \cite{diffusiondrive} and GoalFlow \cite{goalflow}, have shown strong capabilities in generating diverse and high-quality driving trajectories. However, while significant progress has been made in trajectory planning modules, a key bottleneck has not been adequately addressed: the multi-modal fusion architecture that integrates heterogeneous sensor inputs.

Current state-of-the-art methods \cite{goalflow, diffusiondrive, transfuser} mainly rely on transfuser-style architectures for sensor fusion, which simply concatenate image and LiDAR features before applying self-attention mechanisms (Fig. \ref{Fig. 1}a). This approach has two main limitations. First, the conventional histogram-based LiDAR preprocessing discards valuable 3D geometric information by averaging point heights within spatial bins, losing critical cues about object shapes and structures, thereby constraining the performance. Second, standard self-attention mechanisms lack sufficient spatial awareness for processing Bird's Eye View (BEV) data, as they uniformly attend to all positions while ignoring critical spatial priors in driving scenes, such as the greater relevance of forward regions over backward areas and the higher priority of nearby obstacles compared to distant ones. Our experiment result reveals a Contradiction: the multimodal fusion architecture shows only slight performance improvements compared to unimodal architectures, suggesting that transfuser-style architectures treat multi-modal integration as a simple feature combination rather than a structured information integration task.

For these challenges, we propose GMF-Drive, which consists of three components: data preprocessing, perception module, and trajectory planning module. In the first module, GMF-Drive processes raw point clouds into a geometrically-augmented 14-dimensional pillar representation, preserving rich geometric scene information. In the second module, GMF-Drive introduces the GM-Fusion module, which leverages a spatial-aware state-space model (SSM) to achieve linear \(\mathcal{O}(N)\) complexity while maintaining a global receptive field. In the third module, GMF-Drive adopts a truncated diffusion strategy similar to DiffusionDrive, generating plausible trajectories using anchored trajectories.

Our contributions can be summarized as follows:
\begin{itemize}
    \item We design a geometry-enhanced point cloud representation and demonstrate its effectiveness in multi-modal fusion.
    \item We introduce GM-Fusion, a novel fusion architecture powered by a spatial-aware state-space model (BEV-SSM), to achieve superior accuracy over conventional transformers for autonomous driving.
    \item Extensive ablation studies confirm that our key components—geometrically-augmented pillars, BEV-SSM, and hierarchical deformable cross-modal attention—each significantly contributes to our state-of-the-art accuracy on the NAVSIM benchmark.
\end{itemize}
\begin{figure*}[t]
\centering
\includegraphics[width=\textwidth]{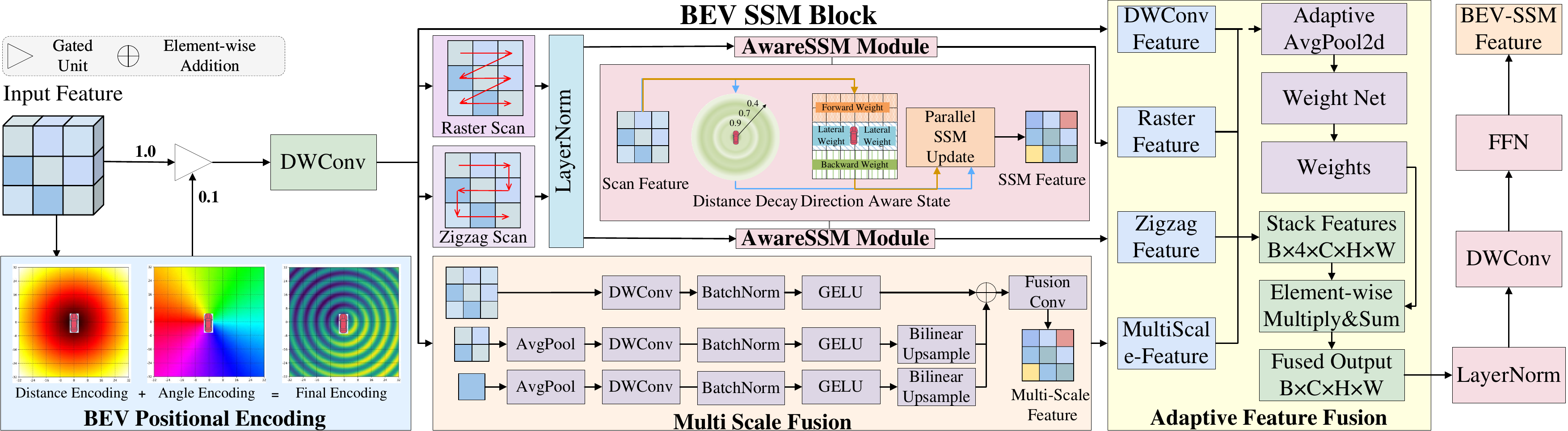} 
\caption{Architecture of the BEV SSM Block. The input features are first processed through two paths: one applies BEV Positional Encoding, while the other maintains the original features. These are combined via gating and then passed through a depthwise separable convolution (DWConv), which performs spatial filtering with reduced parameters by separating depthwise and pointwise operations. The processed features then branch into four parallel paths: (1) a direct path to the adaptive Feature Fusion Module; (2) a path with raster scanning followed by the AwareSSM Module; (3) a path with zigzag scanning followed by the AwareSSM Module; and (4) a path through Multi-Scale Fusion. The Adaptive Feature Fusion Module receives all four inputs and performs learnable weighted fusion rather than simple addition, enabling the model to automatically learn the importance of different serialization strategies for optimal spatial modeling.}
\label{3}
\end{figure*}
\section{Related Works}
\subsection{End-to-End Autonomous Driving}
End-to-end autonomous driving has progressed from early CNN-based methods to advanced multi-modal systems. Initial work like \cite{bojarski2016end} showed CNNs \cite{chua1997cnn} could map images to steering commands, but these methods had limited generalization. Conditional imitation learning improved performance. CILRS \cite{codevilla2018end} used navigational commands to guide driving policies, while LBC \cite{chen2020learning} introduced teacher-student learning with privileged information. The shift to Bird's Eye View (BEV) representations marked a key advance. TransFuser \cite{transfuser} combined image and LiDAR data using transformer, inspiring new BEV methods. UniAD \cite{uniad} integrated perception tasks to improve planning, and VAD \cite{vad} introduced efficient vector representations. Recent work focuses on multi-modal decision-making. SparseDrive \cite{sparsedrive} explored sparse representations, while GoalFlow \cite{goalflow} generated diverse trajectories. However, most methods still rely on heavy transformer architectures, which our work addresses with more efficient models.

\subsection{Multi-Modal Fusion in Autonomous Driving}
Multi-modal fusion in autonomous driving uses three main approaches: early \cite{zhao2024deep,hemker2024healnet}, late \cite{dixit2024deep, shen2025get}, and intermediate fusion \cite{praveen2024recursive,li2024coupled}. Early fusion combines raw sensor data but struggles with different data formats. Late fusion merges high-level decisions but misses cross-modal interactions. Intermediate fusion, now the most common method, uses a transformer to combine features. TransFuser \cite{transfuser} fused image and LiDAR data at multiple scales using attention, outperforming earlier geometry-based methods like ContFuse \cite{liang2018deep}. Recent works like BEVFusion \cite{bevfusion} and FUTR3D \cite{chen2023futr3d} improved fusion in shared feature spaces. However, these methods rely on heavy self-attention, forcing them to use low-resolution features and losing important details. Our work diverges from transformer-centric paradigms by introducing GM-Fusion, which leverages a spatial-aware state-space model to achieve linear complexity while maintaining long-range dependency modeling.

\section{Methodology}
\subsection{Overview}
GMF-Drive (Figure \ref{2}) encodes camera images and a geometrically-rich 14D LiDAR pillar representation using separate ResNet-34 backbones. Across four scales, our GM-Fusion module integrates these modalities via channel attention, a BEV-SSM for linear-complexity spatial modeling, and hierarchical deformable cross-attention (HCA). The resulting features, combined with ego status and trajectory anchors, are then passed to a Diffusion Decoder to generate the final trajectory using truncated diffusion.

\subsection{Geometrically-Augmented Pillar Representation}
Traditional multimodal fusion methods often lose critical geometric information when discretizing LiDAR point clouds into voxels. To address this, we propose a 14-dimensional pillar representation that preserves complete 3D geometric features while maintaining computational efficiency. Given a point cloud $\mathcal{P} = \{p_i\}_{i=1}^N$ where each point $p_i = (x_i, y_i, z_i, r_i, ring_i)$ contains 3D coordinates, reflectance intensity, and ring index, we first discretize the Bird's-Eye View (BEV) space into a regular grid of pillars with resolution $\Delta = 1/\rho$, where $\rho$ represents pixels per meter.

For each pillar at grid location $(u, v)$, we extract a comprehensive feature vector $\mathbf{f}_{uv} \in \mathbb{R}^{14}$ consisting of both pooled point features and statistical descriptors:

Pooled Point Features:
For points $\{p_j\}_{j=1}^{m}$ within pillar $(u,v)$, we compute the pillar center $(x_c, y_c)$ and mean height $z_{\text{mean}}$. The relative offsets are:
\begin{equation}
    \Delta x_{j} = x_{j} - x_{c}, \quad 
    \Delta y_{j} = y_{j} - y_{c}, \quad 
    \Delta z_{j} = z_{j} - z_{\text{mean}}.
\end{equation}
We then apply max-pooling over augmented point features:
\begin{equation}
    \mathbf{f}_{1:8} = \max_{j} \left[ x_{j}, y_{j}, z_{j}, r_{j}, \text{ring}_{j}, \Delta x_{j}, \Delta y_{j}, \Delta z_{j} \right].
\end{equation}

Statistical Features: To capture local geometric structures that histograms fail to represent, we compute intensity statistics including mean and variance of reflectance values, which describe surface scattering properties:
\begin{equation}
    f_{9} = \frac{1}{m}\sum_{j=1}^{m} r_{j},
\end{equation}
\begin{equation}
    f_{10} = \frac{1}{m}\sum_{j=1}^{m}\left(r_{j}-f_{9}\right)^{2}.
\end{equation}

Additionally, we compute four geometric shape descriptors ($f_{11}$ to $f_{14}$) based on principal component analysis PCA \cite{uddin2021pca} of the 3D coordinates. These descriptors characterize the local point distribution and include measures of linearity, planarity, sphericity, and anisotropy. The linearity measure helps detect poles and edges, while planarity is essential for identifying ground surfaces and walls. Sphericity indicates volumetric scatter, and anisotropy captures directional variation to distinguish structured objects from noise.

This 14-dimensional representation preserves critical geometric information typically lost in histogram-based approaches. Unlike methods that only count points, our pillar features encode height variations, intensity patterns, and local surface geometries. This enables our model to better distinguish between different object types with similar point densities but different geometric structures, leading to improved perception accuracy and safer trajectory planning.
\subsection{Gated Mamba Fusion Module}
GM-Fusion integrates multimodal features through three synergistic components. First, gated channel attention aligns and fuses camera and LiDAR features. Second, our BEV-SSM efficiently models spatial dependencies using direction-aware, dual-pattern scanning with distance decay. Third, hierarchical deformable cross-attention (HCA) refines the representation by querying multi-scale image features. This architecture achieves high-precision perception while remaining computationally efficient for high-resolution inputs.
\subsubsection{BEV Positional Encoding.}
Standard positional encoding methods often fail to effectively represent the complex spatial structure of autonomous driving scenarios. To address this limitation, we propose a specialized BEV positional encoding that jointly models distance and directional information relative to the ego vehicle. The method operates in an ego-centric polar coordinate system, where each spatial location is characterized by its Euclidean distance from the ego origin and its angular bearing computed using the two-argument arctangent function. These polar coordinates are transformed into high-dimensional representations through multi-frequency sinusoidal encoding, with both distance and angle features encoded using sine-cosine pairs with exponentially decaying wavelengths.

The key innovation of our approach lies in the dimensional interleaving strategy, where distance and angle encoding components are systematically alternated rather than concatenated as separate blocks. This design forces the joint learning of spatial magnitude and orientation at every feature level, enabling more coherent reasoning about relative positions. By maintaining continuous interaction between distance and directional information, the encoding scheme enhances geometric interpretation and improves perception tasks in autonomous driving. The resulting representation better captures the structured relationships between traffic participants while preserving computational efficiency.
\subsubsection{Dual Scanning Patterns.}
To effectively serialize 2D BEV features for sequential processing, we employ two complementary scanning patterns that capture different spatial relationships:

Raster Scan: This pattern follows a row-by-row traversal, scanning from left to right across each row before moving to the next. In the context of autonomous driving, when aligned with the vehicle's forward direction, raster scanning naturally prioritizes features along the driving path. This makes it particularly effective for capturing long-range dependencies in the direction of travel, which is crucial for anticipating road conditions and distant obstacles.

Zigzag Scan: This pattern follows a serpentine path that alternates direction between rows, ensuring that spatially adjacent locations remain close in the serialized sequence. Unlike raster scanning, which can separate neighboring pixels across row boundaries, zigzag scanning maintains local continuity. This property makes it particularly effective for preserving local spatial relationships and capturing fine-grained geometric patterns near objects.

The complementary nature of these scanning patterns allows our model to capture both global scene structure (through raster scanning) and local geometric details (through zigzag scanning). By processing features through both patterns in parallel and adaptively fusing their outputs, the BEV-SSM block can model multi-scale spatial dependencies essential for comprehensive scene understanding.
\subsubsection{AwareSSM Module.}
The spatial-aware state-space model (AwareSSM) module processes serialized BEV features. It takes scanned features as input and processes them through two parallel branches-distance decay and direction awareness-before feeding them into parallel SSM blocks to generate spatially-informed output features.

The module's directional sensitivity is achieved through learnable state transition matrices that capture the \textit{anisotropic} nature of driving scenarios. We define three learnable transition matrices $\mathbf{A}_{\text{forward}}$, $\mathbf{A}_{\text{lateral}}$, $\mathbf{A}_{\text{backward}} \in \mathbb{R}^{d_{\text{state}}}$ that are dynamically combined based on the scanning pattern according to:
\begin{equation}
    \mathbf{A} = w_{\text{fw}} \mathbf{A}_{\text{forward}} + w_{\text{lat}} \mathbf{A}_{\text{lateral}} + w_{\text{bw}} \mathbf{A}_{\text{backward}},
\end{equation}
where the combination weights $\mathbf{w} = ( w_{\text{fw}}, w_{\text{lat}}, w_{\text{bw}} ) \in \mathbb{R}^3$ are determined by the sequence type. The design strategically allocates directional information processing weights based on operational significance: the forward direction receives the highest initialization weight due to its critical role in safe navigation and collision avoidance, lateral perception employs moderate weighting to monitor lane changes, and the backward direction has the lowest weight while maintaining essential situational awareness.

\begin{table*}[t] 
\centering
\begin{tabular}{@{}lccccccccc@{}} 
\toprule
Method & Input & Img. Backbone & NC $\uparrow$ & DAC $\uparrow$ & TTC $\uparrow$ & Comf. $\uparrow$ & EP $\uparrow$ & PDMS $\uparrow$ \\
\midrule
UniAD~\cite{uniad} & Camera & ResNet-34 & 97.8 & 91.9 & 92.9 & \textbf{100} & 78.8 & 83.4 \\
PARA-Drive~\cite{para-drive} & Camera & ResNet-34 & 97.9 & 92.4 & 93.0 & 99.8 & 79.3 & 84.0 \\
LTF~\cite{transfuser} & Camera & ResNet-34 & 97.4 & 92.8 & 92.4 & \textbf{100} & 79.0 & 83.8 \\
LAW~\cite{law} & Camera & ResNet-34 & 96.4 & 95.4 & 88.7 & \underline{99.9} & 81.7 & 84.6 \\
Transfuser~\cite{transfuser} & C \& L & ResNet-34 & 97.7 & 92.8 & 92.8 & \textbf{100} & 79.2 & 84.0 \\
DRAMA~\cite{drama} & C \& L & ResNet-34 & 98.0 & 93.1 & \textbf{94.8} & \textbf{100} & \underline{80.1} & 85.5 \\
VADv2-$\mathcal{V}_{8192}$~\cite{vadv2} & C \& L & ResNet-34 & 97.2 & 89.1 & 91.6 & \textbf{100} & 76.0 & 80.9 \\
Hydra-MDP-$\mathcal{V}_{8192}$~\cite{hydra} & C \& L & ResNet-34 & 97.9 & 91.7 & 92.9 & \textbf{100} & 77.6 & 83.0 \\
Hydra-MDP-$\mathcal{V}_{8192}$-W-EP~\cite{hydra} & C \& L & ResNet-34 & \textbf{98.3} & 96.0 & 94.6 & \textbf{100} & 78.7 & 86.5 \\
DiffusionDrive \cite{diffusiondrive} & C \& L & ResNet-34 & \underline{98.2} & \underline{96.2} & \underline{94.7} & \textbf{100} & \underline{82.2} & \underline{88.1} \\
\bottomrule
\textbf{GMF-Drive (ours)} & C \& L & ResNet-34 & \underline{98.2} & \textbf{97.3} & 94.2 & \textbf{100} & \textbf{83.3} & \textbf{88.9} \\
\bottomrule
\end{tabular}
\caption{Comparison on planning-oriented NAVSIM \cite{navsim} navtest split with closed-loop metrics. For fair comparison, we uniformly adopt ResNet-34 \cite{resnet} as our feature extraction backbone. “C” \& “L” denotes the use of both Camera and LiDAR as sensor inputs. “$\mathcal{V}_{8192}$” denotes 8192 anchors. “Hydra-MDP-$\mathcal{V}_{8192}$-W-EP” is a variant of Hydra-MDP, which is further trained to fit the EP evaluation metric with additional supervision from the rule-based evaluator and uses weighted confidence post-processing. GMF-Drive simply learns from human demonstrations and infers without post-processing. The \textbf{best} and the \underline{second best} results are denoted by \textbf{bold} and \underline{underline}.}
\label{tab:comparison}
\end{table*}

Complementing this directional awareness, the module implements a distance-based feature adjustment to model the decreasing reliability of distant features. This decay mechanism applies: 
\begin{align}
    \text{decay}_{i} &= \exp\left(-\lambda \cdot d_{i}/d_{\max}\right), \\
    \mathbf{x}_{i}^{\prime} &= \mathbf{x}_{i} \cdot \text{decay}_{i},
\end{align}
where $\mathbf{x}_i$ represents the feature vector at spatial position $i$, $\mathbf{x}'_i$ is the distance-weighted feature, $d_i$ is the distance from the ego vehicle to position $i$, $d_{\text{max}}$ is the maximum distance in the grid, and $\lambda$ is a learnable decay parameter. This ensures nearby objects receive stronger attention while preserving global context awareness.

To accelerate processing while maintaining dependency modeling, the module employs efficient state updates that parallelize traditional sequential transitions. The state calculation utilizes the formulation:
\begin{equation}
\mathbf{u}_t=\mathbf{B}_t \odot \mathbf{C}_t,\qquad
\mathbf{A}_t=\sigma\!\big(\mathbf{A}+\boldsymbol{\Delta}_t\big),
\end{equation}
\begin{equation}
\mathbf{P}_t=\bigodot_{i=1}^{t}\mathbf{A}_i,\qquad
\mathbf{h}_t=\mathbf{P}_t \odot \sum_{j=1}^{t}\left(\mathbf{u}_j \oslash \mathbf{P}_j\right),
\end{equation}
where at step $t$,$B_t$ and $C_t$ are channel-wise linear projections of input; $u_t$ is exogenous drive; $A_t$ represents channel-wise transition gate via sigmoid, governing updating; $P_t$ indicates cumulative retention factor; $h_t$ is parallel-computed hidden state of a first-order, time-varying, channel-wise state-space model, equivalent to gated recurrence.

Overall, the AwareSSM module adaptively adjusts its processing based on scanning patterns (raster or zigzag) and modulates direction weights accordingly. Raster scanning emphasizes forward dependencies using weights $(w_{\text{fw}}, w_{\text{lat}}, w_{\text{bw}})$, while zigzag scanning balances local neighborhood relationships with distinct weights. This approach leverages scanning strategy strengths while maintaining computational efficiency through parallelized operations.
\subsubsection{Adaptive Feature Fusion Module.}

The adaptive feature fusion module dynamically combines features from the BEV-SSM module by analyzing the global scene context. Instead of using fixed averaging, this module automatically learns to adjust the importance weights of different feature sources according to their relevance in the current driving scenario.

The fusion process first extracts global contextual information by aggregating spatial features across the entire input. This global context is then processed through a compact neural network to generate a set of normalized fusion weights. These learned weights are applied to combine the input feature maps from different scanning patterns, producing an adaptively weighted output.

The module dynamically adjusts feature contributions based on driving environment characteristics, with scanning pattern weights optimized according to scene complexity. This adaptive fusion mechanism maintains computational efficiency while ensuring optimal feature representation for varying road conditions.
\section{Experiment}
\subsection{Experimental Setup}
We conducted experiments on the publicly available NAVSIM dataset. To ensure a fair comparison, we adopted the same evaluation metrics as those used in DiffusionDrive. GMF-Drive builds upon DiffusionDrive's codebase, substituting TransFuser modules with our GM-Fusion architecture. We process three concatenated forward-facing camera images ($1024 \times 256$) alongside enhanced pillar-based LiDAR representations with $0.25\,\text{m}$ spatial resolution covering $32\,\text{m}$ longitudinally and $32\,\text{m}$ laterally. Training employs 4 NVIDIA A100 GPUs for 100 epochs with batch size 16, using AdamW optimizer (learning rate $1 \times 10^{-4}$, reduced $10\times$ after epoch 90). The diffusion process uses $T_{\text{trunc}}=50$ during training and requires only 2 denoising steps for inference. Data augmentation is not adopted, and the model outputs 8 waypoints over 4 seconds at $2\,\text{Hz}$.

\begin{table*}[t]
\centering
\begin{tabular}{ccccccccc}
\toprule
Method & Pillar & GM-Fusion & NC $\uparrow$ & DAC $\uparrow$ & TTC $\uparrow$ & Comf. $\uparrow$ & EP $\uparrow$ & PDMS $\uparrow$ \\
\midrule
baseline & & & \textbf{98.20} & 96.20 & \textbf{94.70} & \textbf{100} & 82.20 & 88.10 \\
8d-Pillar & $\surd$ & & 98.10 & 97.27 & 93.64 & \textbf{100} & \textbf{83.43} & 88.61 \\
8d-Pillar+GMF & $\surd$ & $\surd$ & 98.18 & 97.27 & 94.20 & \textbf{100} & 83.26 & 88.82 \\
14d-Pillar+GMF & $\surd$ & $\surd$ & \textbf{98.20} & \textbf{97.34} & 94.19 & \textbf{100} & 83.27 & \textbf{88.85} \\
\bottomrule
\end{tabular}
\caption{Ablation study on system components. Pillar indicates the geometrically-augmented LiDAR representation, and GM-Fusion (GMF) represents our proposed GM-Fusion module. 8-d Pillar represents an 8-dimensional pillar with pooled point features. 14-d Pillar refers to a 14-dimensional pillar with pooled point features and statistical features. The \textbf{best} results are denoted by \textbf{bold}.}
\label{tab:sys_comparison}
\end{table*}

\begin{table*}[t]
\centering
\begin{tabular}{>{\raggedright\arraybackslash}p{3cm}cccccc}
\toprule
Fusion-Style & NC $\uparrow$ & DAC $\uparrow$ & TTC $\uparrow$ & Comf. $\uparrow$ & EP $\uparrow$ & PDMS $\uparrow$ \\
\midrule
baseline (SA) & \textbf{98.20} & 96.20 & \textbf{94.70} & \textbf{100} & 82.20 & 88.10 \\
CA & 98.19 & 96.83 & 94.28 & \textbf{100} & 82.79 & 88.39 \\
C-EffiMamba & 98.05 & 96.56 & 93.92 & \textbf{100} & 82.71 & 88.02 \\
CA+EffiMamba & 98.16 & 96.65 & 94.00 & \textbf{100} & 82.47 & 88.04 \\
HCA+EffiMamba & 98.11 & 97.00 & 93.96 & 99.98 & 82.94 & 88.44 \\
HCA+BEV-SSM & 98.18 & \textbf{97.15} & 94.14 & \textbf{100} & \textbf{83.18} & \textbf{88.69} \\
\bottomrule
\end{tabular}
\caption{Comparison of different fusion architectures. SA: self-attention. CA: cross-attention. C-EffiMamba: cross-efficientmamba \cite{efficientvmamba}. HCA: hierarchical deformable cross-attention. BEV-SSM: our proposed BEV state-space model. The \textbf{best} results are denoted by \textbf{bold}.}
\label{tab:model_comparison}
\end{table*}

\begin{table*}[t]
\centering
\begin{tabular}{lccccccccc}
\toprule
 & HCA & Channel Attention & BEV-SSM & NC $\uparrow$ & DAC $\uparrow$ & TTC $\uparrow$ & Comf. $\uparrow$ & EP $\uparrow$ & PDMS $\uparrow$ \\
\midrule
 & & & & \textbf{98.20} & 96.20 & \textbf{94.70} & \textbf{100} & 82.20 & 88.10 \\
 & $\surd$ & & & 98.09 & 97.03 & 93.87 & 99.98 & 83.03 & 88.46 \\
 & $\surd$ & $\surd$ & & 98.23 & 97.13 & 94.14 & \textbf{100} & 83.12 & 88.67 \\
 & $\surd$ & & $\surd$ & 98.18 & 97.15 & 94.14 & \textbf{100} & 83.18 & 88.69 \\
 & $\surd$ & $\surd$ & $\surd$ & \textbf{98.20} & \textbf{97.34} & 94.19 & \textbf{100} & \textbf{83.27} & \textbf{88.85} \\
\bottomrule
\end{tabular}
\caption{Ablation study on GM-Fusion internal components. HCA: hierarchical deformable cross-attention. Channel Attention: cross-modal channel alignment. BEV-SSM: our proposed BEV state-space model. The \textbf{best} results are denoted by \textbf{bold}.}
\label{tab:hierarchical_attention}
\end{table*}
\subsection{Quantitative Comparison}
As shown in Table \ref{tab:comparison}, GMF-Drive sets a new state-of-the-art on the NAVSIM navtest split with a PDMS score of 88.9, outperforming the previous best, DiffusionDrive, by 0.8 points. This gain is achieved using the same ResNet-34 backbone and sensor inputs, demonstrating the superiority of our GM-Fusion architecture over traditional transformer-based fusion.

The strengths of our approach are evident in key sub-metrics. GMF-Drive leads in drivable area compliance (DAC) with a score of 97.3 (+1.1 vs. DiffusionDrive), validating our hypothesis that preserving fine-grained spatial features improves scene understanding. It also achieves the top ego progress (EP) score of 83.3, proving that efficient fusion enhances both safety and driving efficiency.

Furthermore, GMF-Drive significantly surpasses vocabulary-based methods. It beats a heavily-tuned Hydra-MDP variant by 2.3 PDMS points, despite that method relying on extra rule-based supervision and post-processing. This highlights the advantage of our generative fusion approach over discrete trajectory selection.

While most methods achieve near-perfect Comfort scores, the primary differentiators are collision avoidance (NC), DAC, and EP, where GMF-Drive consistently improves. A marginal decrease in TTC is substantially offset by gains in other critical metrics, resulting in superior overall performance.

\subsection{Ablation Study}
\subsubsection{Overall System Component Analysis.}
The baseline DiffusionDrive model is progressively enhanced through several key innovations, as demonstrated in Table \ref{tab:sys_comparison}. Each modification contributes to measurable performance improvements in the evaluation metrics.

First, replacing the baseline LiDAR encoding with our 8d-pillar representation alone boosted the PDMS score from 88.10 to 88.61. This gain was driven by significant improvements in DAC (96.52→97.27) and EP (82.54→83.43), confirming that richer geometric inputs enhance perception even with a standard fusion module. Next, integrating our Gated Mamba Fusion (GMF) module further lifted the PDMS to 88.82. Despite a minor dip in EP, the overall improvement demonstrates that our spatially-aware fusion mechanism effectively leverages these geometric features. Finally, employing the full 14d-pillar representation yielded the top PDMS score of 88.85. While the gain over the 8d version was modest (+0.03), it brought consistent, small improvements across all metrics. This suggests the 8d representation captures the most critical geometric cues, while the additional statistical features provide marginal refinement.

\begin{figure*}[t]
\centering
\includegraphics[width=\textwidth]{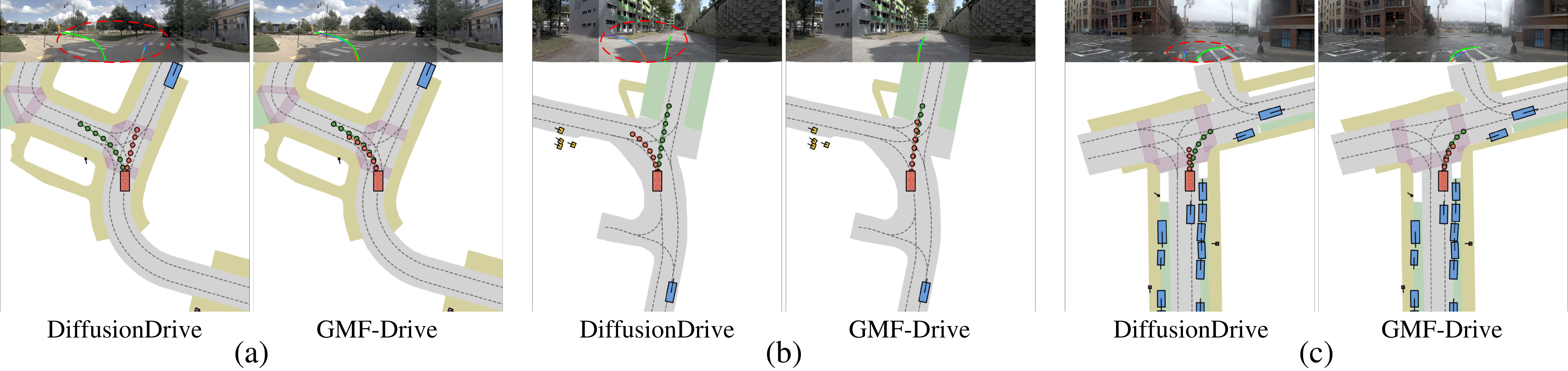} 
\caption{Qualitative comparison of DiffusionDrive and GMF-Drive on representative scenarios from NAVSIM navtest split. (a) Left-turn scenario. (b) Straight-line intersection traversal. (c) Right-turn scenario. For each scenario, the left subfigure displays DiffusionDrive outputs and the right subfigure shows GMF-Drive (Ours) results. The green trajectory indicates the ground truth and the other trajectory represents the model-generated trajectory.}
\label{4}
\end{figure*}

Collectively, these components deliver a 0.75-point PDMS improvement over the baseline, demonstrating that both our data representation and fusion architecture are critical for achieving state-of-the-art performance.

\subsubsection{Fusion Architecture Design Analysis.}

The systematic evaluation of fusion strategies presented in Table \ref{tab:model_comparison} provides compelling evidence for our architectural decisions. Beginning with the baseline self-attention (SA) mechanism, we observe that transitioning to cross-attention (CA) yields a measurable improvement in performance, with PDMS increasing from 88.10 to 88.39. This enhancement stems from the fundamental advantage of cross-attention in establishing direct feature correspondences between camera and LiDAR modalities, enabling more effective cross-modal interaction compared to the concatenation-based fusion used in the baseline.

A particularly revealing finding emerges from the C-EffiMamba variant, which attempts to replace the transformer architecture entirely with a general-purpose state-space model. The performance degradation to 88.02 PDMS demonstrates that naive application of standard sequential models fails to capture the complex spatial relationships inherent in driving scenarios. This result strongly suggests that effective perception systems require specialized architectures incorporating appropriate inductive biases for the autonomous driving domain.

The hybrid architectures present an interesting progression in performance. While the basic combination of cross-attention and EfficientMamba (CA+EffiMamba) shows only marginal improvement to 88.04 PDMS, the hierarchical variant (HCA+EffiMamba) achieves significantly better results at 88.44 PDMS. This performance gap highlights the critical importance of multi-scale processing in sensor fusion, where hierarchical attention mechanisms can adaptively focus on relevant regions across different spatial resolutions. The hierarchical architecture's ability to maintain contextual awareness while processing fine details appears crucial for accurate perception.

The most significant advancement comes with the HCA+BEV-SSM configuration, which incorporates our specialized BEV-optimized state-space model. Achieving 88.69 PDMS, this architecture demonstrates how task-specific modifications to the SSM can overcome the limitations observed in general-purpose implementations. The BEV-SSM's directional scanning patterns and spatial priors prove particularly effective for processing Bird's-Eye View (BEV) representations, leading to superior performance in trajectory prediction and obstacle detection tasks. These results collectively validate our GM-Fusion approach, which strategically combines the strengths of hierarchical attention mechanisms and domain-optimized sequential modeling for autonomous driving perception.
\subsubsection{GM-Fusion Component Analysis.}
Table \ref{tab:hierarchical_attention} quantifies the contribution of each GM-Fusion component. Adding only hierarchical deformable cross-attention (HCA) boosts the baseline PDMS from 88.10 to 88.46, driven by gains in DAC and EP. This validates HCA's effectiveness in capturing multi-scale, cross-modal interactions.

Combining HCA with Channel Attention further lifts the PDMS to 88.67, primarily by improving NC and TTC. This highlights the importance of explicitly aligning heterogeneous camera and LiDAR features before fusion. Interestingly, pairing HCA with our BEV-SSM achieves a similar PDMS of 88.69. This demonstrates that BEV-SSM's spatial modeling can be as effective as channel-wise alignment for fusion, while offering superior computational efficiency due to its linear complexity. Finally, the complete GM-Fusion module, integrating all three components, achieves the peak PDMS score of 88.85. The final performance gain confirms that Channel Attention and BEV-SSM are complementary: one aligns features across modalities, while the other models spatial dependencies within the fused representation. This synergy is key to our model's state-of-the-art performance.

\subsection{Qualitative Comparison}
To visually demonstrate the advantages of our approach, we present comparative trajectory visualizations in Figure \ref{4}. In the left-turn scenario(a), DiffusionDrive generates a straight trajectory diverging from the ground truth, exhibiting directional inconsistency. During straight-line traversal (b), the baseline method produces unexpected leftward deviation. For the right-turn scenario (c), DiffusionDrive maintains a straight trajectory misaligned with the required geometry. These patterns indicate limitations in spatial reasoning within transformer-based fusion frameworks.

Conversely, GMF-Drive achieves accurate path alignment across all scenarios. Our method correctly navigates left-turns (a) and right-turns (c) while maintaining stable straight-line motion (b). This performance directly stems from our architectural innovations: the geometrically-augmented pillar representation preserves essential 3D structural information, while the GM-Fusion enables spatially-aware feature integration through directional scanning sequences. By adaptively prioritizing forward regions during multimodal processing, our approach generates trajectories that better respect spatial constraints compared to transformer-based fusion.
\section{Conclusion}
In this work, we present GMF-Drive, a novel end-to-end autonomous driving framework that integrates geometrically-augmented pillar representation with a gated spatial-aware state-space model for multimodal fusion in perception systems. This fusion approach effectively replaces current transformer-based fusion frameworks. Through comprehensive experiments on the NAVSIM benchmark, our ablation studies validate the rationality of the fusion architecture design, while quantitative results demonstrate that GMF-Drive achieves state-of-the-art performance.

\bibliography{aaai2026}

\end{document}